\documentclass[10pt, logo, copyright]{nvidiatechreport}

\usepackage{mdframed}
\usepackage[utf8]{inputenc} %
\usepackage[T1]{fontenc}    %

\usepackage{amsfonts}       %
\usepackage{nicefrac}       %
\usepackage{microtype}      %

\usepackage[numbers]{natbib}
\usepackage{tabto}
\usepackage{xspace}
\usepackage{adjustbox}
\usepackage{wrapfig}
\usepackage{dblfloatfix}
\usepackage{footmisc}
\usepackage{float}
\usepackage{capt-of}
\usepackage{listings}
\definecolor{codegreen}{HTML}{2E7D32}
\definecolor{codepurple}{HTML}{7B1FA2}
\definecolor{codegray}{HTML}{757575}
\definecolor{codeblue}{HTML}{1565C0}
\definecolor{codeorange}{HTML}{E65100}
\definecolor{codebg}{HTML}{FAFAFA}
\lstdefinestyle{python}{
  language=Python,
  basicstyle=\ttfamily\small,
  keywordstyle=\color{codepurple}\bfseries,
  stringstyle=\color{codegreen},
  commentstyle=\color{codegray}\itshape,
  emphstyle=\color{codeblue},
  emph={load,do,narrate,lm,extract,convert,smart_convert,invoke_tool},
  emph={[2]ctx,d_type},
  emphstyle={[2]\color{codeorange}},
  backgroundcolor=\color{codebg},
  numbers=left,
  numberstyle=\tiny\color{codegray},
  numbersep=6pt,
  frame=single,
  rulecolor=\color{codegray},
  columns=fullflexible,
  breaklines=true,
  showstringspaces=false,
  tabsize=4,
  postbreak=\mbox{\textcolor{red}{$\hookrightarrow$}\space},
}

\usepackage{natbib}
\usepackage{hyperref}
\usepackage{url}
\usepackage{booktabs}
\usepackage{multirow}
\usepackage{tabularx}
\usepackage{arydshln}

\usepackage{colortbl}
\usepackage{amsmath}
\usepackage{makecell}
\usepackage{hhline}
\usepackage{array}
\usepackage{diagbox}
\usepackage{graphicx}

\usepackage{fp} %

\usepackage{bbm}
\usepackage{amsmath}
\usepackage{amssymb}
\usepackage{fontawesome}
\usepackage{pifont}%
\usepackage{MnSymbol}
\usepackage[ruled,lined]{algorithm2e}
\DontPrintSemicolon
\SetKwInOut{Intermediate}{Intermediate}

\usepackage{booktabs}
\usepackage{multirow}
\usepackage{makecell}
\usepackage{rotating}
\definecolor{periwinkle}{HTML}{7977B8}
\definecolor{hlparam}{HTML}{FFF3B0}      %
\definecolor{hlderived}{HTML}{D0E8FF}    %
\newcommand{\hlp}[1]{{\setlength{\fboxsep}{0.5pt}\colorbox{hlparam}{\ensuremath{\mathstrut #1}}}}

\newcommand{\hld}[1]{{\setlength{\fboxsep}{0.5pt}\colorbox{hlderived}{\ensuremath{\mathstrut #1}}}}

\usepackage{hyperref}
\usepackage{url}
\usepackage{titlesec}
\usepackage[capitalize,noabbrev]{cleveref} %
\usepackage{enumitem}

\crefname{failure}{failure mode}{failure modes}
\crefname{gissue}{guardrail shortcoming}{guadrail shortcomings}
\crefname{aview}{alternative view}{alternative views}
\crefname{argument}{argument}{arguments}
\crefname{cargument}{counter-argument}{counter-arguments}
\crefname{step}{step}{steps}
\crefname{barrier}{barrier}{barriers}
\crefname{wdefinition}{definition}{definitions}
\newlist{assumptions}{enumerate}{1}
\setlist[assumptions]{label=\textbf{A\arabic*}}
\crefname{assumptionsi}{assumption}{assumptions}
\Crefname{assumptionsi}{Assumption}{Assumptions}

\newlist{observations}{enumerate}{1}
\setlist[observations]{label=\textbf{O\arabic*}}
\crefname{observationsi}{observation}{observations}
\Crefname{observationsi}{Observation}{Observations}

\title{CPU-ARC: Agentic Reasoning Caching for on-CPU Reasoning Reuse}

\author{
Peter Belcak, Greg Heinrich, Saurav Muralidharan, Shizhe Diao, Pavlo Molchanov
}

\titlespacing*{\paragraph}
  {0pt}   %
  {0pt}   %
  {6pt}   %

\makeatletter
\def\gcmidrule{\arrayrulecolor{lightgray}%
    \noalign{\ifnum0=`}\fi
    \@ifnextchar[{\@gcmidrule}{\@gcmidrule[\cmidrulewidth]}}
\def\@gcmidrule[#1]{\@ifnextchar({\@@gcmidrule[#1]}{\@@gcmidrule[#1]()}}
\def\@@gcmidrule[#1](#2)#3{\@@@gcmidrule[#3]{#1}{#2}}
\def\@@@gcmidrule[#1-#2]#3#4{\global\@cmidla#1\relax
    \global\advance\@cmidla\m@ne
    \ifnum\@cmidla>0\global\let\@gtempa\@cmidrulea\else
    \global\let\@gtempa\@cmidruleb\fi
    \global\@cmidlb#2\relax
    \global\advance\@cmidlb-\@cmidla
    \global\@thisrulewidth=#3
    \@setrulekerning{#4}
    \ifnum\@lastruleclass=\z@\vskip \aboverulesep\fi
    \ifnum0=`{\fi}\@gtempa
    \noalign{\ifnum0=`}\fi\futurenonspacelet\@tempa\@xgcmidrule}
\def\@xgcmidrule{%
   \ifx\@tempa\gcmidrule
       \vskip-\@thisrulewidth
       \global\@lastruleclass=\@ne
   \else \ifx\@tempa\morecmidrules
       \vskip \cmidrulesep
       \global\@lastruleclass=\@ne\else
       \vskip \belowrulesep
       \global\@lastruleclass=\z@
   \fi\fi
   \ifnum0=`{\fi}
  \arrayrulecolor{black}}%
\makeatother

\begin{abstract}
\end{abstract}

\begin{document}
\maketitle

\textbf{Abstract.}
Large language models re-invent the wheel all the time.
They conflate task-solving with the execution of the solutions, which forces them to solve every problem from scratch.
We show that once a task has been reasoned through, the application of its solution to a new problem instance can be done very efficiently and with little involvement from language models.
We introduce agentic reasoning caching, a method for constructing robust CPU-executable agentic reasoning caches (ARCs) from LLM reasoning traces.
ARCs show on-par performance with frontier LLMs across a range of difficult mathematical and scientific tasks, are fully explainable, continually trainable, robust to input perturbations, and 16-291x cheaper to invoke than the source model.

\vspace{0.4em}
\noindent\textbf{Links:}\enspace
\href{https://nv-dler.github.io}{\textcolor{OliveGreen}{Lab}}\enspace\textbar\enspace
\href{https://research.nvidia.com/labs/lpr/lfi/}{\textcolor{OliveGreen}{Project Moonshot: Local Frontier Intelligence}}

\section{Introduction}
\label{section:introduction}

Language models solve every problem from scratch, no matter how common.
However, many of their solutions apply not just to the exact problems posed but extend to entire classes of problems in which the individual problem instances vary in their phrasing or some of their parameters.
\Cref{table:math500-perturbation-example} illustrates this on an example drawn from a popular mathematical benchmark.

\begin{table}[h!]
\centering
\footnotesize
\begin{tabularx}{\linewidth}{@{} l X X X @{}}
\toprule
& \textbf{Original} & \textbf{Rephrased} & \textbf{Reparameterized} \\
\midrule
\makecell[lt]{\textbf{Question}}
& How many positive whole-number divisors does \hlp{196} have?
& Find how many positive integers evenly divide \hlp{196}.
& How many positive whole-number divisors does \hlp{144} have? \\
\addlinespace[4pt]
\hdashline
\addlinespace[4pt]
\makecell[lt]{\textbf{Reasoning}}
& \multicolumn{2}{c}{\parbox[t]{\dimexpr(\linewidth-13\tabcolsep)/3\relax}{%
I need to find the prime factorisation of \hlp{196} first. $\hlp{196} = \hld{4} \times \hld{49} = \hld{2^{2}}\times \hld{7^{2}}$. \newline\rule{0pt}{2.2ex}For a number with prime factorisation $p_{1}^{a_{1}} \times p_{2}^{a_{2}} \times \cdots$, the number of divisors is $(a_{1}+1)(a_{2}+1)\cdots$. \newline\rule{0pt}{2.2ex}For $\hlp{196} = \hld{2^{2}}\times \hld{7^{2}}$: $\hld{(2+1)(2+1)}=\hld{3\times 3}=\hld{\mathbf{9}}$.%
}}
& I need to find the prime factorisation of \hlp{144} first. $\hlp{144} = \hld{16} \times \hld{9} = \hld{2^{4}}\times \hld{3^{2}}$. \newline\rule{0pt}{2.2ex}For a number with prime factorisation $p_{1}^{a_{1}} \times p_{2}^{a_{2}} \times \cdots$, the number of divisors is $(a_{1}+1)(a_{2}+1)\cdots$. \newline\rule{0pt}{2.2ex}For $\hlp{144} = \hld{2^{4}}\times \hld{3^{2}}$: $\hld{(4+1)(2+1)}=\hld{5\times 3}=\hld{\mathbf{15}}$. \\
\bottomrule
\end{tabularx}
\caption{A MATH-500 problem, its rephrasing while keeping all the parameters the same, its reparametrization while keeping the phrasing the same, and the corresponding reasoning traces. The rephrased variant changes only the wording; the reparameterized variant substitutes the numerical parameter and adjusts the arithmetic accordingly. Note how neither the rephrasing nor the reparametrization change even a single word in the reasoning steps leading to the final answer, not to mention their substance.}
\label{table:math500-perturbation-example}
\end{table}

AI agents and AI-enabled software more broadly often rely on repetitive queries to LLMs featuring the same instructions and plans but different payloads.
The LLMs comprehending the same prompts and subsequently reasoning through their tasks from scratch for each new query often results in a significant amount of work being done repeatedly across separate inference instances.

System-level caching (SLC) methods aim to address such inefficiencies by recognizing repetitive patterns in LM conversations and caching them at the level of the serving system as a whole rather than within any single inference instance. Maintained at the system level, an SLC's entries are shared across all of the system's inference instances and remain reusable beyond the instance in which they were born.
SLC methods usually bring one or more of the following benefits:
\begin{itemize}
  \item \textbf{Cost and latency reduction.} SLC methods can reduce the cost and/or the latency of LLM calls by re-using partial or complete results of earlier LLM calls.

  \item \textbf{Improved correctness.} SLCs can be reviewed asynchronously by humans or a committee of judge agents to ensure that the responses or steps of the most popular queries are correct.

  \item \textbf{Improved explainability or interpretability.} SLCs can be inspected by humans to understand the reasoning process and the final answer or result, providing explanation or at the very least an interpretation of responses to popular queries.
\end{itemize}

When choosing or designing an SLC method, the following considerations are important:
\begin{itemize}
  \item \textbf{Local robustness.} How well does the SLC method cover the class of queries it is supposed to cover? How robust is the SLC method to small perturbations of the queries?
  \item \textbf{Correctness upon hit.} If the cache is hit, is the result correct? How much confidence can one have in the correctness of the cached result given a hit from the query neighbourhood it covers?
  \item \textbf{Explainability or interpretability.} How much can one learn by inspecting the caches? How good are the guarantees of correctness provided by the cache entries? Can they be inspected by an authority (e.g., a regulatory body) to provide a guarantee on behavior?
  \item \textbf{Cost.} How expensive is it to build, maintain, and use the SLC?
\end{itemize}

Existing work on SLC falls into two broad categories:
\begin{itemize}
  \item \textbf{Monolithic caching.} Methods such as ephemeral prefix caching \cite{kwon2023efficient}, programmatic conversation caching \cite{openai2026responses,google2026gemini_cache}, and semantic query matching \cite{bang2023gptcache} look up transcripts of previous conversations on the basis of input prefix overlaps, programmatically supplied conversation IDs, and semantic similarities between different queries.
  They store entire conversations and their corresponding intermediate generation states (e.g., KV caches or Mamba hidden states) and aim to re-use the largest possible contiguous portion of the query's generation states wherever possible.
  Monolithic caches are usually not robust to even the slightest perturbations of the conversation, and when they are (e.g., when employing semantic similarity search), it is often at the expense of their correctness.
  Monolithic caches of responses can be inspected to guarantee absolute correctness.
  They are usually compute-cheap to build, maintain, and use but are memory- and bandwidth-expensive to maintain and use.
  
  \item \textbf{Modular caching.} Instead of storing the entire conversational transcripts and their generation artifacts, modular SLC methods first perform a mechanical form of decomposition of the existing contexts and their generation states or a distillation of the context into critical logical steps \cite{singh2026semantic,gupta2026reasoncache,wang2026crystal}.
  Later, when encountering an apparently similar problem, the decomposed artifacts are re-supplied to the language model to accelerate their reasoning process.
  Only parts of the response are cached, and even in the case of a perfect cache hit, a full new generation (though accelerated) of the response is still required.
  This makes their coverage more robust to parametric perturbations of the inputs but makes it impossible to guarantee correctness upon review.
  Modular caches sometimes offer interpretable insights into the reasoning process of the underlying models.
  They are considerably more compute-expensive to build and use than monolithic caches but tend to be more memory- and bandwidth-efficient to maintain and use.
\end{itemize}

We introduce agentic reasoning caching, a (see below) hybrid SLC method that produces agentic reasoning caches (ARCs) that are robust in coverage, can be asynchronously ensured to be correct, are fully explainable, and are cheap to build, maintain, and use.
We call them \emph{agentic reasoning caches} because their elements are small, specialized fractional agents that can be re-used on new parameters.
Our approach rests on two observations:
\begin{observations}
  \item\label{observation:execution_is_cheap}
  Once a particular problem has been reasoned through and the solution has been found, the execution of the solution -- the computation or data manipulation behind the problem -- can be performed with little involvement from language models.
  In other words, the search for a solution can be decoupled from the execution of the solution found.
  \item\label{observation:reasoning_can_be_reused}
  Much of LM's reasoning about a particular problem instance can be reused for a set of problems extending beyond the singleton instance.
\end{observations}

Leveraging \Cref{observation:execution_is_cheap}, we choose to convert LM conversation transcripts to code, and allow that code to make use of only small language models to perform semantic similarity search and elementary natural language processing tasks such as value extraction, format conversion, classification, or summarization, and only where needed.
Leveraging \Cref{observation:reasoning_can_be_reused}, we choose to parametrize the code in a way that allows it to be re-used for all problems that are \textit{equirational} to the problem solved in the original transcript (i.e., can be solved using the same reasoning).
Further, we choose to produce code in such a manner that does not rely on fixed variable, parameter, and function names but on their natural language descriptions and examples, thus allowing for a liberal re-use of all logical components of previously cached logic in all semantically equivalent scenarios, no matter how phrased, structured, or parameterized.

Agentic reasoning caching is a hybrid of monolithic and modular SLC because it decomposes the logic of an LLM conversation into a network of modules -- \textit{semantic functions} -- that can be re-used individually and also provides their composition into a single function that can be used to solve the problem cached.
Because of the semantic functions' wide parametrization, ARCs are robust to logically valid parametric perturbations of the problem.
Since every ARC entry is stored as code, ARCs are fully explainable and correctible upon human or AI inspection.
Since ARCs are stored as code using only SLMs, they are cheap to maintain and use since they rely mostly on CPU execution.
They are further relatively cheap to build, coming at 12-84x the cost of the original reasoning.

\begin{wrapfigure}[12]{r}{0.38\textwidth}
\vspace{-12pt}
\begin{lstlisting}[style=python, numbers=none]
def serve(request):
  try:
    result = cache.use(request)
    if result is not False:
      return result
  except CacheInvocationError:
    pass
  return llm(request)
\end{lstlisting}
\vspace{-6pt}
\caption{Gated use of an agentic reasoning cache. The cache is tried first; on a miss or error, the request falls through to the full LLM.}
\label{figure:intelligence_cache_setup}
\label{figure:intelligence_cache_minimal_code_example}
\vspace{-12pt}
\end{wrapfigure}
Agentic reasoning caching is particularly suitable for agentic systems that repeatedly use the same prompts with variable payloads.
In an agentic system, using an ARC can be attempted before forwarding the request to the cloud LLM, allowing for a significant reduction in the latency and cost in the case of a cache hit.
Owing to their reliance on simple CPU execution and SLMs, agentic reasoning caches are so resource-efficient that they can be brought to edge devices, reducing the cost further.
\Cref{figure:intelligence_cache_setup} briefly illustrates the gated use of an agentic reasoning cache in an AI system where the cache re-use is attempted before forwarding the request to the full LLM.

Our contributions are:
\begin{enumerate}[label=\textbf{C\arabic*}]
  \item The introduction of agentic reasoning caching, a hybrid SLC method that is robust in its coverage, can be asynchronously ensured to be correct by human or AI inspection, is fully explainable, and is cheap to build, maintain, and use.
  \item The evaluation of agentic reasoning caching on a range of challenging mathematical and science benchmarks, including its success rates, correctness, robustness, and cost-saving capabilities.
  \item A brief discussion of the intended placement and implications of this work.
\end{enumerate}

Our long-term goal is to bring ARC to real production agents to help lower their cost. In this initial, exploratory work, we deliberately focus on the hardest case: rather than agentic workloads of varied difficulty, we test whether ARCs can match frontier-level model performance on difficult mathematical and scientific benchmarks that are commonly used to measure intelligence. Succeeding at that (as we show we do) indicates that the method is not capped by the ability of models to produce these small ARCs, and that, by extension, the broader class of agentic tasks currently driven by frontier-level models is plausibly cacheable as well.

\section{Agentic reasoning caching}
\label{section:intelligence_caching}

Let $\mathcal{S},\mathcal{B},\mathcal{T}$ be the source, builder, and target language model, respectively, and $\mathcal{A}$ be an answering oracle that answers all queries correctly.
Here, the source $\mathcal{S}$ is the capable, expensive model whose intelligence we aim to cache; the builder $\mathcal{B}$ is the model that analyzes $\mathcal{S}$'s reasoning transcripts and constructs cache entries; and the target $\mathcal{T}$ is the small, cost-efficient model on which caches are to execute at inference time.
Additionally, in some places we make use of LM judges $\mathcal{J}$, which are simply language models tasked with verifying the correctness and faithfulness of intermediate and final outputs throughout construction and evaluation.
Let $q(\bullet_1,\bullet_2,\dotsc,\bullet_n)$ be a query template with $n$ parameters.

Let $q(o)$ be a query instance submitted for caching with $o$ being a specific value vector for $q$'s parameters. 

Assume that $\mathcal{A}(q(o))$ exists, meaning that $q(o)$ is solvable.
Observe that there exists a set of parameter vectors $\Pi$ such that for all $\pi \in \Pi$, $q(\pi)$ is equirational to $q(o)$.
This is because at the very least, $q(o)$ is equirational to itself.

Agentic reasoning caching is based on the following two assumptions:

\begin{assumptions}
  \item\label{assumption:solvability} For every $q(o)$ that is solvable there exists a program $p(\bullet_1,\bullet_2,\dotsc,\bullet_n)$ with access to an SLM such that $p(\pi) = \mathcal{A}(q(\pi))$ for all $\pi \in \Pi$.
  \item\label{assumption:constructibility} $p(\bullet_1,\bullet_2,\dotsc,\bullet_n)$ can be constructed by $\mathcal{B}$ from knowing $q(o)$ and $\mathcal{S}(q(o))$.
\end{assumptions}

\Cref{assumption:solvability} effectively amounts to the belief that every solvable problem is solvable with a combination of code and a small language model in a way that extends to all equirational variants of the problem.
\Cref{assumption:constructibility} expresses the belief that it is possible to encode into $p$ by some language model $\mathcal{B}$ the intelligence in $\mathcal{S}$ required to solve $q(o)$ and its equirational variants using $q(o)$ and its output under $\mathcal{S}$ as starting points.

Agentic reasoning caching is the embodiment of the above assumptions.
Implementing \Cref{assumption:solvability}, we ask the builder model to analyze source model conversation transcripts to be cached to find which parts (together with their ranges of validity) of their inputs can be varied without requiring substantive changes to the reasoning.
Further implementing \Cref{assumption:constructibility}, we ask the builder model to comprehend the source transcript and construct a program that can solve the problem posed in the original query considering the above-found input parts as parameters that can be varied.
To ensure maximal robustness and cache coverage, we insist that the program is a rich composition of independently invokable units representing logical steps that constitute the reasoning process, which we call semantic functions.
An agentic reasoning cache is then a set of semantic functions combined with a many-to-one map of problem instances to the natural language descriptions of the semantic functions that can solve the respective instances.

The remainder of this section describes the semantic functions that constitute programs $p$ (\Cref{subsection:semantic_functions}) and the cache construction (\Cref{subsection:cache_construction}) and invocation (\Cref{subsection:cache_invocation}) processes.

\subsection{Semantic functions}
\label{subsection:semantic_functions}

A \emph{semantic function} is an executable abstraction representing one logical step of the source reasoning trace, designed to be maximally reusable across semantically equivalent contexts.
They are the matter of which agentic reasoning caching is made up.
In our implementation, it is simply a Python callable that takes a \texttt{Context} object and returns a dictionary mapping natural-language descriptions to values:
\[
  f\colon \texttt{Context} \;\longrightarrow\; \texttt{dict[str, Any]}.
\]
Unlike conventional functions whose inputs, outputs, and callees are referenced by fixed identifiers, semantic functions reference everything by natural-language descriptions and examples.
This enables semantic discovery: a caller need not know the callee's name, only a description of what it does, and a callee need not know the names of its inputs, only their descriptions.
This design allows any component of a previously cached computation to be reused in any semantically equivalent scenario regardless of how it is phrased or structured (\Cref{fig:semantic_function_example}).
Further, if a change (such as a generalization introducing new parameter) is made to a dependency semantic function, none of its callers need to be updated.
Instead, its new, generalized description will open it to a larger set of callers, again without affecting its past callers as long as the new description semantically entails the old one.

\paragraph{Runtime primitives.}
Semantic functions are designed to use a small set of runtime primitives that mediate all data flow and computation:
\begin{itemize}[nosep]
  \item \texttt{load(ctx, query)}: retrieve a value from the layered context memory via vector similarity search. An optional \texttt{d\_type} parameter triggers type coercion on the retrieved value.
  \item \texttt{do(ctx, query, payloads)}: discover and invoke another semantic function by natural-language description. Pushes an isolated memory layer pre-populated with the given payloads, executes the matched function, and returns its output dictionary.
  \item \texttt{narrate(ctx, message)}: record a human-readable reasoning step in the narration trace, providing a running explanation of the computation for later inspection.
  \item \texttt{lm(ctx, prompt)}: query the target model $\mathcal{T}$ for an elementary natural-language processing task such as classification, summarization, or short-form generation.
  \item \texttt{extract(ctx, text, prompt)}: use $\mathcal{T}$ to perform verbatim value extraction from a piece of text. Supports \texttt{d\_type} for type coercion on the extracted value.
  \item \texttt{convert(ctx, value, type)} / \texttt{smart\_convert(ctx, text, type)}: multi-tier type coercion. Programmatic conversion is attempted first; if it fails, \texttt{smart\_convert} asks $\mathcal{T}$ to rewrite the value into a Python literal that is then validated by \texttt{ast.literal\_eval()}.
  \item \texttt{invoke\_tool(ctx, name, args)}: call a registered external tool (e.g., terminal or a web search API), enabling semantic functions to integrate with tool-augmented pipelines.
\end{itemize}

\paragraph{Type safety.}
Because values flow between functions through semantic search rather than typed variable bindings, type coercion is critical.
Both \texttt{load} and \texttt{extract} accept an optional \texttt{d\_type} parameter that coerces the retrieved or extracted value to the desired Python type.
When direct coercion fails, \texttt{smart\_convert} invokes $\mathcal{T}$ to rewrite the value into a Python literal, which is then parsed via \texttt{ast.literal\_eval()} to ensure syntactic validity and type correctness.

\paragraph{Composition via \texttt{do} and \texttt{load}.}
Semantic functions are composed into networks through the \texttt{do()} primitive.
The \texttt{do()} primitive invokes a semantic function by describing its intention imperatively in natural language and passing parameters annotated by their natural language descriptions.
A caller passes described parameter values into \texttt{do()}, which pushes a fresh, isolated memory layer onto the context stack, populates it with the payloads, and dispatches to the callee found by semantic search over the function registry.
The callee accesses its inputs via \texttt{load()} and stores its outputs into the same layer.
Upon return, the caller retrieves results from the callee's output dictionary with \texttt{load(result, query)}.
This mechanism enables hierarchical decomposition: a root function can delegate to arbitrarily nested sub-functions based on its intention rather than their named identities, with each sub-function operating in its own scope while sharing the underlying context stack.
See \Cref{fig:example_sf_geom} in Appendix~\ref{appendix:worked_example_geom} for a complete cache entry illustrating these primitives on an ideal gas law problem from SciBench.

\paragraph{Context and memory model.}
The \texttt{Context} object maintains a stack of memory layers, each backed by a CPU-powered vector store using a small semantic embedding model.
When \texttt{load()} is called, it searches the layers top-down, returning the best match from the most local scope.
Each invocation of \texttt{do()} pushes a new layer and pops it upon completion, providing natural scoping that prevents unintended interference between sibling calls.

\begin{figure}[t]
\begin{lstlisting}[style=python]
def smart_function(ctx):
    narrate(ctx, "Checking if a valid rotation exists that "
            "maps blue vertices to red positions")

    vertex_set = load(ctx, "the blue vertex set B", d_type=list)
    modulus = load(ctx, "the vertex count n", d_type=int)

    B = set(vertex_set)

    # Compute difference set via a sub-function
    diff_result = do(ctx,
        "Compute the difference set D(B) for the vertex "
        "subset B with modulus n",
        payloads={
            "the set of vertices B": list(B),
            "the modulus n for modular arithmetic": modulus,
        })

    diff_set = load(diff_result, "difference set",
        d_type=set, ctx=ctx)

    # A valid rotation k exists iff k is not in the
    # difference set D(B)
    all_rotations = set(range(modulus))
    valid_k_values = all_rotations - diff_set

    narrate(ctx, f"Valid rotations: {sorted(valid_k_values)}")

    return {
        "valid rotation exists": len(valid_k_values) > 0,
        "valid k values": sorted(valid_k_values),
    }
\end{lstlisting}
\caption{A semantic function that checks whether a set of blue vertices on a regular $n$-gon can be rotated so that every blue vertex lands on a previously red position.
It loads its inputs from the context via \texttt{load()}, delegates the difference-set computation to another semantic function via \texttt{do()}, performs set arithmetic to find valid rotations, and records its reasoning via \texttt{narrate()}.
The return value is a dictionary with natural-language keys, following the semantic function contract.}
\label{fig:semantic_function_example}
\end{figure}

\subsection{Cache}
\label{subsection:cache}

\subsubsection{Construction}
\label{subsection:cache_construction}

Given a query instance $q(o)$ and the source model's transcript $\mathcal{S}(q(o))$, the builder model $\mathcal{B}$ constructs the program $p$ through five phases: input--output isolation, decomposition, parameter recognition, planning, and implementation.

\paragraph{Phase 1: Input-output isolation.}
$\mathcal{B}$ extracts the user input from the query and isolates the expected output (the final answer or result) from the transcript $\mathcal{S}(q(o))$.
This separation grounds the subsequent phases: the user input drives parameter identification in Phase~3, while the expected output serves as the correctness reference against which candidate implementations are tested in Phase~5.

\paragraph{Phase 2: Decomposition.}
$\mathcal{B}$ analyzes the transcript $\mathcal{S}(q(o))$ and decomposes the reasoning into a hierarchical list of logical operations.
Each operation is described imperatively, naming the basic individual inputs it consumes (values, entities, relationships, quantities) rather than referring to the problem as a monolithic description.
The target granularity is function-sized units of logic: chunks small enough to be implemented as individual semantic functions (\Cref{subsection:semantic_functions}), yet large enough to represent a coherent reasoning step.
During decomposition, $\mathcal{B}$ also identifies reusable sub-computations, each representing an operation likely to recur across equirational variants or entirely different queries, and marks them for extraction into standalone semantic functions.

\paragraph{Phase 3: Parameter recognition.}
Using both the user input and the decomposition as context, $\mathcal{B}$ identifies the class of problems the query belongs to and determines which parts of the input are the natural input parameters that vary across instances of that class.
$\mathcal{B}$ produces a generic input template $q(\bullet_1, \bullet_2, \dotsc, \bullet_n)$ and extracts the concrete value of each parameter from the original input, paired with a descriptive name and format examples.
Every recognized parameter must be a basic, self-contained piece of information that the reasoning directly consumes (a number, a name, a formula, a relationship, a list of items, a table, and so on) and never an entire problem description or a multi-sentence narrative.
Parameters that are naturally variable-size collections (lists, tuples, tables) are kept as single parameters rather than split into individual elements.
The output of this phase is the generic input template $q(\bullet_1, \bullet_2, \dotsc, \bullet_n)$ together with a dictionary mapping parameter descriptions to their concrete values in the original instance.
The phase concludes with an analysis of the logical validity ranges of the individual parameters.

\paragraph{Phase 4: Planning.}
Given the transcript, the decomposition, the recognized parameters, and any tools that were available to $\mathcal{S}$ during the original reasoning, $\mathcal{B}$ produces a structural plan for the program.
The plan specifies which semantic functions to create, their dependency structure, the order in which they should be implemented (bottom-up, so that each function's dependencies are available before it is written), and which function serves as the root.
A configurable granularity setting (fine, medium, or coarse; specified by the cache user) controls how aggressively small operations are merged into larger functions: at the fine end, nearly every decomposition step becomes its own function, while at the coarse end only the major reasoning stages are preserved.
The plan is designed for robustness to re-parameterization: it ensures that the resulting program structure does not depend on the specific parameter values $o$ of the original instance.

\paragraph{Phase 5: Implementation.}
\label{subsection:cache_testing}
$\mathcal{B}$ enters an agentic loop in which it iteratively writes candidate semantic functions.
A function remains a candidate until it passes all checks and is registered in the cache. It may be rewritten multiple times before it reaches that point.

For each candidate function, $\mathcal{B}$ first searches the existing registry for semantically equivalent functions and inspects their source code to assess whether they can be reused directly or adapted.
It then writes the candidate function body using the runtime primitives described in \Cref{subsection:semantic_functions}.
The candidate is validated syntactically and via static analysis of narration quality, checking that narration calls are present and that their number is commensurate with the function's control-flow complexity.
$\mathcal{B}$ then tests the candidate by dispatching it with the original parameters and comparing its output against the corresponding fragment of the transcript, validated by a judge $\mathcal{J}$.
On success, the function is registered in the cache.
On failure, $\mathcal{B}$ receives structured feedback (error messages, judge critiques, stack traces) and rewrites the candidate accordingly.

Once all individual functions are registered, $\mathcal{B}$ tests the root function end-to-end in two stages.
In the first stage (correctness), the root function is dispatched with the original parameters $o$ and its output is judged by $\mathcal{J}$ against both the transcript and the golden label.
In the second stage (extraction robustness), a real invocation is simulated: $\mathcal{T}$ extracts parameters from the raw user input, the root function is dispatched with the extracted parameters, and the output is judged by $\mathcal{J}$.
This extraction round-trip is repeated $N$ times to test robustness to variance in the extraction process.

On completion, the root function and its entry-point descriptions are registered in the cache's vector-indexed registries.
Both the generic input template $q(\bullet_1, \bullet_2, \dotsc, \bullet_n)$ and the original user input $q(o)$ are registered as entry points, enabling retrieval by semantic similarity at invocation time.

\subsubsection{Invocation}
\label{subsection:cache_invocation}

Once a cache has been constructed, serving a new query $q(\pi)$ for $\pi \in \Pi$ proceeds through four steps.

\begin{enumerate}[label=\textbf{IS\arabic*}]
  \item \textbf{Entry-point search.}
  The user input is embedded and compared against the entry-point registry via cosine similarity.
  If the best result similarity exceeds a configurable threshold, the corresponding root function description is returned. Otherwise, the query is forwarded to the source model $\mathcal{S}$ as a cache miss.

  \item \textbf{Parameter value extraction.}
  The extractor LM, built on $\mathcal{T}$, extracts each expected parameter from the user input.
  The extractor is parameterized by the parameter descriptions stored alongside the root function during registration, ensuring that the extraction is aligned with the function's expected inputs.

  \item \textbf{Dispatch.}
  A fresh \texttt{Context} is created and populated with the extracted parameter values.
  The root function is executed, potentially invoking sub-functions via \texttt{do()}, querying $\mathcal{T}$ via \texttt{lm()}, and recording reasoning via \texttt{narrate()}.

  \item \textbf{Output.}
  The function returns a dictionary of named results.
  The full narration trace, accumulated during execution, serves as a human-readable explanation of how the answer was derived.
  Note that while the narrations produced by semantic functions might eventually end up appearing formulated in a somewhat structured and machine-like manner, it would not be too difficult to employ a small transcription model that would make them sound more human-like if genuinely desired by the end users.
\end{enumerate}

\medskip
\noindent
A key property of this design is that invocation uses only the target (small) model $\mathcal{T}$ and CPU execution; the expensive source model $\mathcal{S}$ and builder model $\mathcal{B}$ are required only at construction time.
This asymmetry is what makes agentic reasoning caching economical at scale: the one-time construction cost is amortized over arbitrarily many invocations, each of which runs at the cost of a few $\mathcal{T}$ calls and deterministic code execution.

\subsubsection{Compilation}
\label{subsection:cache_compilation}

Although invocation already replaces $\mathcal{S}$ with the cheaper $\mathcal{T}$, it still relies on vector-similarity search at every \texttt{do()} and \texttt{load()} call to resolve function names and parameter bindings at runtime.
Once a cache has matured (e.g., its function registry is stable and no further entries are expected), these dynamic lookups can be resolved statically.
ARCs can be compiled using static resolution by traversing the semantic function network, replacing each \texttt{do()} dispatch with a direct Python function call to the resolved callee, and replacing each \texttt{load()} with a direct dictionary lookup on the caller's payload, producing a self-contained Python module that requires no embedding engine and no vector store at execution time.

The benefit of this is twofold.
First, compiled caches execute faster because every semantic search is eliminated and replaced by $O(1)$ name-based dispatch, reducing invocation latency to that of pure Python execution plus any remaining \texttt{lm()} calls to $\mathcal{T}$.
Second, the compiled artifact is a readable, standalone Python file that can be inspected, versioned, and deployed independently of the caching runtime.
See \Cref{fig:example_compiled_geom} for a compiled version of the cache entry in \Cref{fig:example_sf_geom}.
The trade-off is that compilation freezes the function network: any subsequent cache update (e.g.\ adding a new function or generalizing a parameter) requires recompilation.
Compilation is therefore best applied as a finalization step after the cache has grown sufficiently and been validated as useful, rather than during iterative construction.

\subsubsection{Choice of $\mathcal{T}$}
\label{subsection:target_model_choice}

The ultimate invocation cost of an agentic reasoning cache with the design as outlined in this section depends to a large extent on the target model $\mathcal{T}$ that executes the cached functions at invocation time.
How small can $\mathcal{T}$ be?
The precise answer depends on the domain and the operational range of the cache: caches that must handle diverse, open-ended inputs would likely place larger demands on the language model than those that operate over narrower domains.
In our experiments, the smallest frontier-lab models: Claude Haiku 4.5, Gemini 3.0 Flash Lite, and GPT 5.4 Nano always sufficed as $\mathcal{T}$, delivering correct results as a part of ARCs at a fraction of $\mathcal{S}$'s cost.

To explore how much further the target model can be shrunk, we conducted a fine-tuning experiment.
We collected 20\% of the real usage examples from our benchmark evaluations, used $\mathcal{S}$ to produce a synthetic training set roughly ten times larger, and finetuned Llama~3.2~1B~\cite{meta2024llama32} on the resulting data.
The finetuned 1B-parameter model proved sufficient for the target role, and we began using it as $\mathcal{T}$ in subsequent runs.
\Cref{appendix:target_model_production} describes the data collection, synthetic augmentation, and training procedure in detail.

The iterative construction process naturally tolerates an imperfect $\mathcal{T}$: because the builder observes invocation outputs and can adjust the language model calls in semantic functions accordingly, transient errors from a weaker target model are corrected over successive rounds rather than propagated silently.
We believe a 1B-parameter target is adequate for the majority of agentic reasoning caches, and future work could push the boundary further to models with only hundreds of millions of parameters, such as those in the IBM Granite 4 family~\cite{ibm2025granite4}.

We further note that in situations when the cache is being built in the background of LLM inference before it is deployed for use, it is actually possible to build the cache with a more powerful target model first, and then collect all logs of invocations of the more powerful target model, finetune the smallest possible target model that works to replace the more powerful target, and only then deploy the cache live, resulting in greater efficiency.

\section{Evaluation}
\label{section:evaluation}

We evaluate agentic reasoning caches along three axes: correctness, perturbation robustness, and cost.
We first define the properties of experimental interest, then describe the benchmarks and the experimental setup, and finally showcase the results.
Appendix~\ref{appendix:worked_example_geom} walks through a complete ideal gas law example from SciBench, showing the question (\Cref{fig:example_question_geom}), source reasoning (\Cref{fig:example_transcript_geom}), semantic functions and their compiled form, and cache output (\Cref{fig:example_output_geom}).

\subsection{Measured properties}
\label{subsection:measured_properties}

\paragraph{Correctness.}
To evaluate how well the logic of cache entries aligns with the benchmark ground truth and source model reasoning, we measure two quantities.
Firstly, we measure \textit{golden label alignment}: the fraction of benchmark problems for which the cache produces the correct answer as determined by the benchmark's golden label.
Secondly, we measure \textit{transcript answer alignment}: the fraction of problems for which LM judges assess the cache's answer as semantically equivalent to the source model's answer.
The former captures absolute performance; the latter captures fidelity to $\mathcal{S}$.
Close alignment between $\mathcal{S}$ and the cache on both metrics would indicate that the cache faithfully reproduces the source model's reasoning and as such is a good replacement for the intelligence of the source model on the problems given.
A large gap in which the cache underperforms $\mathcal{S}$ against the golden label would signal that the cache construction has lost information.
Note that because the source model's reasoning can be at odds with golden labels at times, looking at both of these measured qualities in conjunction with the source model's performance on that benchmark allows us to analyze whether the model favors the ground truth golden labels or the source model's reasoning when building the cache entries.

\paragraph{Perturbation robustness.}
Agentic reasoning caches are designed to generalize beyond the singleton problem instance used during construction despite seeing only the original query $q(o)$ and the source model's reasoning trace $\mathcal{S}(q(o))$.
We evaluate this generalization through three progressively stronger perturbation testing modes, building on the terminology introduced in \Cref{section:intelligence_caching}:
\textit{phrasal} (equiparametric) testing, which varies only the wording of the question;
\textit{parametric} (equiphrasal) testing, which varies only the parameter values; and
\textit{combined} (equisubstantive) testing, which varies both simultaneously.
The generation and judging procedures for each mode are described in \Cref{subsection:perturbation_testing}.

\paragraph{Cost.}
We report three cost figures per benchmark problem, all computed from retail API pricing (including prompt-cache write and read rates where applicable):
\begin{itemize}
  \item \textit{Source cost}: the cost of the original $\mathcal{S}$ solve that produces the transcript to be cached.
  \item \textit{Builder cost}: the cost of constructing the cache entry from the transcript using $\mathcal{B}$.
  \item \textit{Target cost}: the cost of invoking the cache on a query, comprising only $\mathcal{T}$ calls and deterministic execution.
\end{itemize}
The ratio of target cost to source cost quantifies the per-invocation savings; the builder cost divided by the source cost quantifies the one-time construction overhead.

\subsection{Perturbation testing}
\label{subsection:perturbation_testing}

Perturbation testing is applied only to benchmark problems for which the cache already produces the correct answer according to the golden label.
In other words, to be considered for robustness testing, cache entries need not agree with the source model's transcript answer, only with the ground truth.
This selection criterion ensures that the substance being perturbed is known to be correctly handled by the cache, so that any degradation observed under perturbation can be attributed to a lack of robustness rather than to a pre-existing correctness failure.
Foreshadowing the results, a secondary benefit is that the set of golden-matching samples is, in our evaluation, typically larger than the set of transcript-matching samples, yielding more test material.

Every perturbation mode follows a three-stage pipeline.
Let $\mathcal{G}$ denote the generator model that produces perturbation candidates (in our experiments, $\mathcal{G} = \mathcal{S}$), $\mathcal{J}_G$ the generation judges that validate candidate perturbations, and $\mathcal{J}_T \subset \mathcal{J}$ the testing judges that evaluate cache outputs against ground truth golden labels or golden-passing narrations.
\begin{enumerate}[label=\textbf{PT\arabic*}]
  \item \textbf{Candidate generation.} $\mathcal{G}$ produces a set of candidate perturbations of the original problem $q(o)$: rephrasings, reparameterizations, or both, depending on the mode.
  \item \textbf{Validation judging.} The generation judges $\mathcal{J}_G$ evaluate each candidate to verify that it is a faithful perturbation rather than an artifact of hallucination or reasoning error by $\mathcal{G}$.
  Candidates that fail any judge are discarded. The specific validation criteria differ between phrasal and parametric perturbations and are described below.
  \item \textbf{Result evaluation.} Each accepted candidate is run through the cache. The testing judges $\mathcal{J}_T$ then evaluate the cache's output against the appropriate ground truth for that candidate.
\end{enumerate}

\paragraph{Phrasal perturbation (equiparametric).}
\textit{Generation.}
Given a problem $q(o)$ for which the cache is correct, $\mathcal{G}$ produces candidate rephrasings of $q(o)$ while holding all parameter values in $o$ fixed.
Each candidate must express the substantively same question in sufficiently different phrasing.

\textit{Validation.}
$\mathcal{J}_G$ evaluate each candidate against three criteria: (i)~all parameter values from $o$ are preserved exactly, (ii)~the phrasing is sufficiently different from the original, and (iii)~the substance of the question is unchanged.
Only candidates that pass all judges in $\mathcal{J}_G$ unanimously are accepted.
This filters out rephrasings that inadvertently alter parameter values or the meaning of the question due to generator hallucination.

\textit{Evaluation.}
The cache is invoked on each accepted variant and its output is evaluated by $\mathcal{J}_T$ against the golden label: because the parameters have not changed, the correct answer is the same as for the original problem instance.

\paragraph{Parametric perturbation (equiphrasal).}
\textit{Generation.}
Given a problem $q(o)$ for which the cache is correct, $\mathcal{G}$ decomposes the source reasoning into logical steps, identifies the parameters and their valid ranges and constraints within $\Pi$, and generates candidate parameter vectors $\pi' \neq o$ that respect all identified constraints.
The phrasing template $q(\bullet)$ is held fixed throughout.
Alongside each candidate, $\mathcal{G}$ produces two narrations: the original narration, which traces the solution of $q(o)$ and arrives at the known golden answer, and the final narration, which traces the solution of $q(\pi')$ under the new parameter values and arrives at a proposed new answer.
\textit{Validation.}
$\mathcal{J}_G$ verify that (i)~the question structure is preserved, (ii)~at least one parameter value has actually changed, (iii)~all domain constraints are satisfied, (iv)~the original narration correctly reproduces the known golden answer for $q(o)$, and (v)~the final narration is logically consistent and arrives at a coherent answer for $q(\pi')$.
Criterion~(iv) serves as a calibration check: if $\mathcal{G}$ cannot correctly re-derive the known answer for the original instance, its reasoning about the perturbed instance is not trustworthy, and the candidate is discarded.
This filters out reparameterizations that violate domain constraints or contain reasoning errors introduced by $\mathcal{G}$.
\textit{Evaluation.}
Because parameter values have changed, the golden label of the original instance no longer applies.
The cache is invoked on each accepted variant $q(\pi')$ and its output is compared by $\mathcal{J}_T$ against the answer from the \textit{final narration}, which serves as the ground truth for the perturbed instance.

\paragraph{Combined perturbation (equisubstantive).}
Combined testing is a two-stage pipeline that applies both perturbation modes in sequence.
Stage~1 generates, validates, and evaluates phrasal variants as described above.
Stage~2 takes each accepted phrasal variant and applies parametric perturbation, with its own generation, validation, and evaluation done independently of the first stage.
The result is a set of variants that differ from $q(o)$ in both phrasing and parameter values.
Each stage has its own independent judging committee.
Combined testing is the strongest robustness evaluation: it exercises the cache's ability to handle novel formulations of novel instances simultaneously.

\subsection{Benchmarks}
\label{subsection:benchmarks}

We evaluate on six benchmarks spanning competition mathematics, graduate-level science, grade-school reasoning, and multidisciplinary knowledge.

\begin{itemize}
  \item \textbf{MATH-500.} A set of 500 competition-style mathematics problems drawn from high-school and undergraduate competitions~\cite{hendrycks2021measuring}. Problems range from algebra and number theory to geometry and combinatorics.
  \item \textbf{AIME 2024, 2025, 2026.} Problems from the American Invitational Mathematics Examination across three consecutive years~\cite{MAA2024AIME,MAA2025AIME,MAA2026AIME}. AIME problems are significantly harder than MATH-500, requiring multi-step reasoning and creative problem solving.
  \item \textbf{GPQA Diamond.} The Diamond subset of the graduate-level Google-Proof Q\&A (GPQA) benchmark~\cite{rein2024gpqa}, comprising expert-written science questions in physics, chemistry, and biology that are designed to be unsearchable and require graduate-level domain knowledge.
  \item \textbf{GSM8K.} A benchmark of approximately 8,000 grade-school math word problems~\cite{cobbe2021training} that test multi-step arithmetic reasoning. Despite the elementary subject matter, the benchmark is useful for evaluating whether cache construction preserves basic reasoning chains.
  \item \textbf{MMLU-Pro.} An extended version of the Massive Multitask Language Understanding benchmark~\cite{wang2024mmlu} with harder, reasoning-focused questions across diverse academic disciplines.
  \item \textbf{SciBench.} A collection of science problems drawn from university-level textbooks~\cite{wang2023scibench} in physics, chemistry, and mathematics, requiring both domain knowledge and quantitative reasoning.
\end{itemize}

We evaluate on mathematical and science benchmarks not because we intend the agentic reasoning caches to be used on such problems, but because these problems are generally considered to be difficult enough to be used to assess model intelligence.
The rationale of this evaluation is then that if our agentic reasoning caching methodology can successfully cache the intelligence required to solve these problems, and these are commonly used proxies to model intelligence, then by extension of maxima, it should be able to cache all intelligence in the appropriate domains.

\subsection{Experimental setup}
\label{subsection:experimental_setup}

\paragraph{Models.}
The source model $\mathcal{S}$ and the builder model $\mathcal{B}$ are both Claude Opus 4.5 with extended thinking enabled at a budget of 4096 tokens.
The target model $\mathcal{T}$ was first Claude Haiku 4.5 and then Llama 3.2 1B as described in \Cref{subsection:target_model_choice}, both being small, cost efficient, and run at temperature $0.0$.

\paragraph{Embedding and retrieval.}
Semantic search during cache invocation (\Cref{subsection:cache_invocation}) uses Embedding Gemma~\cite{vera2025embeddinggemma} (a roughly 300M-parameter model) to produce 768-dimensional text embeddings for all \texttt{do()} and \texttt{load()} dispatch, indexed via a FAISS flat index for CPU-based similarity lookup.
Compiled caches (\Cref{subsection:cache_compilation}) bypass the embedding engine entirely.

\paragraph{Perturbation generation.}
The generator $\mathcal{G}$ is Claude Opus 4.5, identical to $\mathcal{S}$.
The generation judges $\mathcal{J}_G$ are a committee of 2 Claude Opus 4.5 instances and the testing judges $\mathcal{J}_T$ are a committee of 2 Claude Opus 4.5 instances, each with extended thinking enabled.
Acceptance requires unanimity within each committee.

\paragraph{Cache construction.}
Construction proceeds per-sample: each query instance $q(o)$ is built in an isolated, self-contained cache, which simplifies debugging and error attribution.
Immediately upon successful construction, the per-sample cache is merged into a master cache that accumulates the entries of all processed samples.
The master cache is the artifact that serves queries at invocation time; the per-sample caches are retained only as diagnostic artifacts.

\subsection{Results}
\label{subsection:results}

\begin{table*}[t]
\centering
\footnotesize
\setlength{\tabcolsep}{3pt}
\caption{Results across benchmarks.
\textit{S\,Acc}: source model accuracy (transcript vs.\ golden label);
\textit{C\,Acc}: cache accuracy (cache vs.\ golden label);
\textit{Fidelity}: cache--source agreement (cache vs.\ transcript).
Perturbation columns report pass rates for phrasal, parametric, and combined iso-substantive rewrites.
Cost columns show average per-sample cost at retail API prices;
\textit{Build\,$\times$}\,=\,builder\,/\,source cost;
\textit{Savings\,$\times$}\,=\,source\,/\,cache cost.
See \Cref{appendix:cost_breakdowns} for per-role token usage.}
\label{tab:main_results}
\begin{tabular}{l ccc ccc ccc cc}
\toprule
& \multicolumn{3}{c}{Correctness (\%)} & \multicolumn{3}{c}{Perturbation (\%)} & \multicolumn{3}{c}{Avg.\ Cost (\$/sample)} & \multicolumn{2}{c}{Multipliers} \\
\cmidrule(lr){2-4} \cmidrule(lr){5-7} \cmidrule(lr){8-10} \cmidrule(lr){11-12}
Benchmark & S\,Acc & C\,Acc & Fidelity & Phrasal & Param. & Combined & Source & Builder & Cache & Build$\times$ & Savings$\times$ \\
\midrule
MATH-500            & 98.6  & 98.8  & 99.6  & 96.7  & 94.5  & 91.6  & 0.0388 & 1.1818 & 0.0004 & 30.5  & 88.7  \\
AIME~'24            & 100.0 & 100.0 & 100.0 & 96.0  & 99.3  & 95.1  & 0.1101 & 1.4198 & 0.0004 & 12.9  & 262.4 \\
AIME~'25            & 83.3  & 93.3  & 90.0  & 98.4  & 91.2  & 93.4  & 0.1203 & 1.4517 & 0.0004 & 12.1  & 291.1 \\
AIME~'26            & 90.0  & 96.7  & 90.0  & 97.9  & 90.9  & 90.8  & 0.1197 & 1.5477 & 0.0004 & 12.9  & 281.4 \\
GPQA Diamond        & 83.8  & 83.8  & 98.5  & 93.8  & 93.9  & 94.0 & 0.0708 & 1.7781 & 0.0019 & 25.1  & 38.1  \\
GSM8K               & 98.5  & 98.0  & 99.5  & 96.2  & 94.2  & 92.5  & 0.0101 & 0.8481 & 0.0006 & 83.7  & 16.3  \\
MMLU-Pro            & 88.0  & 89.0  & 100.0 & 95.7  & 96.7  & 91.9  & 0.0345 & 1.4084 & 0.0015 & 40.8  & 22.9  \\
SciBench            & 85.2  & 88.2  & 99.4  & 97.5  & 98.8  & 96.4  & 0.0434 & 1.2572 & 0.0005 & 29.0  & 83.0  \\
\bottomrule
\end{tabular}
\end{table*}

All costs in \Cref{tab:main_results} use March 2026 API rates for cost accounting.
The source and builder models (Claude Opus 4.5) are priced at \$5.00/M input tokens, \$25.00/M output tokens, \$6.25/M cache-write tokens, and \$0.50/M cache-read tokens.
The builder's multi-turn implementation loop is prompt-cached, so the majority of its input tokens are served from cache at the cache-read rate, significantly reducing builder costs relative to a na\"ive implementation without caching.
For cost accounting, calls to the finetuned Llama 3.2 1B model in the target, extractor, and converter roles are valued at \$0.023/M input tokens and \$0.194/M output tokens, with no prompt-cache discount.
The ``cache cost'' column is the sum of target, extractor, and converter costs incurred at invocation time.

\subsection{Discussion}
\label{subsection:discussion}

\paragraph{Correctness.}
We observe that cache accuracy (C\,Acc) matches or exceeds source accuracy (S\,Acc) on seven of the eight benchmarks.
The improvement is most pronounced on the competition mathematics problems of AIME, where the cache achieves 93.3\% vs.\ the source's 83.3\% on AIME~'25 and 96.7\% vs.\ 90.0\% on AIME~'26.
We attribute this to cases where the source model reasons correctly and its reasoning is therefore used to produce a correct cache entry, but a small computational or logical error in the final steps leads the source model to an imprecise answer, whereas the cache executes the same reasoning as deterministic code and avoids such slips.
The sole exception is GSM8K, where the cache (98.0\%) trails the source (98.5\%) by a narrow margin.
Fidelity (the rate at which the cache agrees with the source transcript) is consistently high, exceeding 98\% on six of eight benchmarks and reaching 100\% on MMLU-Pro.
Even where fidelity is lower (AIME~'25 and~'26 at 90.0\%), the cache retains the ability to produce golden-correct answers when the transcript answer is itself incorrect.

\paragraph{Robustness.}
Across all benchmarks and perturbation modes, cache entries handle at least 90\% of perturbations correctly (the lowest observed rate is 90.8\% for combined perturbations on AIME~'26).
Phrasal perturbation rates match or exceed parametric rates on four of the eight benchmarks, indicating that the parameter extraction and reasoning logic are generally more robust to surface-level reformulations than to changes in the underlying parameter values.
Combined perturbation testing is the most demanding mode and yields the lowest rates in six of eight benchmarks, as expected.

\paragraph{Cost.}
We note that builder costs range from $12\times$ to $84\times$ the source invocation cost.
This multiplier is characteristically lower for difficult problems where the source model's reasoning is long but the resulting computation is straightforward to execute (e.g.\ $12$--$13\times$ for AIME), and higher for simpler problems where the source model responds briefly but the cache still requires full construction (e.g.\ $84\times$ for GSM8K).
Conversely, the savings multiplier at invocation time ranges from $16\times$ to $291\times$, with considerably higher savings on difficult competition mathematics (AIME: $262$--$291\times$) where the source model's per-query cost is highest.
Even on simple benchmarks such as GSM8K, the cache delivers a $16\times$ cost reduction per invocation.
In absolute terms, cache invocation costs are uniformly low across all benchmarks (\$0.0004--\$0.0019 per sample), owing to the use of a small finetuned open-source target model.
Let $B$ denote the one-time builder cost, $S$ the source cost per query, and $C$ the cache cost per invocation.
For $S > C$, the break-even point is $B/(S-C)$ invocations, rounded up to the next whole invocation; the Build\,$\times$ multiplier $B/S$ approximates this when $C$ is negligible relative to $S$.
Using the reported cost multipliers, ARCs for AIME problems pay for themselves after approximately 13 neighbourhood invocations, while GSM8K requires approximately 90.

\section{Intended placement and implications}
\label{section:placement_and_implications}

The results of \Cref{section:evaluation} can be interpreted as follows: at a two-digit cost multiplier increase, the logic behind mathematical and scientific reasoning of language models can be converted into a form that, when invoked, reduces the cost by a two-digit multiplier.
It would be prudent to ask what situations benefit the most from this approach and what the forward-looking implications are.

We imagine future caching work in this direction to be placed in agentic systems and AI-enabled software in which the same reasoning or the same set of instructions is frequently applied to varying payloads. Based on our familiarity with the field, such configurations are common in enterprise agentic software and AI-enabled software more broadly.
We see value in installing agentic reasoning caches onto endpoints that handle repetitively equi-instructed payloads accumulating usage over hours or weeks, and then applying the methodology outlined in \Cref{section:intelligence_caching} to produce a more efficient version of that endpoint, with the presence of a large language model fallback in case the incoming queries are sufficiently novel or the cache invocation results in a failure.

Even as currently implemented, agentic reasoning caching can learn continuously: a cache that has been built can be continuously added to without having to rebuild the existing entries from scratch, as would be the case for finetuned language models. Moreover, the agentic reasoning cache can be inspected for correctness and is resolutely cheaper than the use of language models that, while smaller, would still have to be sufficiently large to handle the complex reasoning about the payload.

Going forward, we note that the phenomenon of semantic function reuse, as enabled by the search-based function discovery available during the construction stage (\Cref{subsection:cache_construction}) and the natural-language intent-style descriptions of semantic functions and their parameters, enables a great deal of reuse once a cache grows sufficiently large.
While the problem of generalizing multiple similar but not identical semantic functions into one coherent, more general semantic function is not addressed by this work, we believe that beyond a certain point agentic reasoning caches could overtake a large proportion of workloads from large language models simply by densely populating the domains of most intense interest by their users while growing at a sublinear rate in proportion to the number of input samples.
This would be applicable to domains that do not require a great deal of creativity and are instead focused on the execution of particular instructions.

It is beyond the scope of this work, but we foresee that applying agentic reasoning caching to post-training data corpora could lead to the development of a new class of models -- semantic models -- that live above neural models such as SLMs/LLMs and are fully explainable, highly efficient, and continually trainable.
In such a configuration, neural models, which currently stand at or near the top of the intelligence stack, become a support layer between hardware and intelligent semantic software -- ``neuralware.''

The fact that our present work demonstrates that frontier-level intelligence can, on problems of interest, be cached to rely only on a single 1B-parameter model in a phrasally and parametrically robust manner gives us strong hope for the ability to perform the entire class of non-creative intelligence tasks on the edge in the future.

\clearpage
{
  \small
  \bibliographystyle{unsrt}
  \bibliography{bibliography}
}

\clearpage
\appendix

\section{Worked example: ideal gas law}
\label{appendix:worked_example_geom}

This appendix walks through a single benchmark problem end-to-end, showing every artifact the agentic reasoning caching system produces: the original question, the source model's reasoning transcript, the semantic functions that constitute the cache entry, their compiled form, and the final cache output with its narration trace.
The example is an ideal gas law problem from SciBench~\citep{wang2023scibench} (predicting the pressure of ethane) with three semantic functions exercising temperature conversion, pressure unit conversion, and the root ideal-gas-law computation.
Both the source model's answer and the cache output match the golden label ($\approx 5130\,\mathrm{kPa}$).
This example was selected because it exercises all core mechanisms (parameter extraction via \texttt{load()}, sub-function delegation via \texttt{do()}, result retrieval via \texttt{load(result,\,\ldots)}, reasoning narration via \texttt{narrate()}, value extraction via \texttt{extract()}, and target-model querying via \texttt{lm()}) while requiring an LLM call at invocation time to classify the temperature unit, demonstrating that \texttt{lm()} and \texttt{extract()} survive compilation.

\begin{figure}[H]
\begin{mdframed}[backgroundcolor=codebg, linecolor=codegray]
Suppose that $10.0 \mathrm{~mol}\;\mathrm{C}_2\mathrm{H}_6\mathrm{(g)}$ is confined to $4.860 \mathrm{~dm}^3$ at $27^{\circ}\mathrm{C}$. Predict the pressure exerted by the ethane from the perfect gas.
\end{mdframed}
\caption{Original benchmark question from SciBench (ideal gas law).}
\label{fig:example_question_geom}
\end{figure}

\begin{figure}[H]
\begin{mdframed}[backgroundcolor=codebg, linecolor=codegray]
{\small
\textbf{Predicting pressure using the perfect gas law}

\medskip
\textbf{Given information.}\quad
Amount of ethane: $n = 10.0\;\mathrm{mol}\;\mathrm{C_2H_6}$.
Volume: $V = 4.860\;\mathrm{dm}^3$.
Temperature: $T = 27^{\circ}\mathrm{C} = 300\;\mathrm{K}$.

\medskip
\textbf{Ideal gas equation.}\quad
For a perfect (ideal) gas, $PV = nRT$.
Rearranging for pressure: $P = \frac{nRT}{V}$.

\medskip
\textbf{Substitution.}\quad
Using $R = 8.314\;\mathrm{kPa{\cdot}dm^3{\cdot}mol^{-1}{\cdot}K^{-1}}$:
\[
P = \frac{(10.0)(8.314)(300)}{4.860}
  = \frac{24\,942}{4.860}
  \approx \boxed{5.13 \times 10^3\;\mathrm{kPa} = 5.13\;\mathrm{MPa}}.
\]
This is approximately $51\;\mathrm{bar}$ or $50.6\;\mathrm{atm}$.
}
\end{mdframed}
\caption{Source model $\mathcal{S}$ reasoning transcript for the question in \Cref{fig:example_question_geom}. The answer $\approx 5130\;\mathrm{kPa}$ matches the golden label ($\approx 50.7\;\mathrm{atm}$).}
\label{fig:example_transcript_geom}
\end{figure}

\begin{figure}[H]
\begin{lstlisting}[style=python, basicstyle=\ttfamily\footnotesize, title={\footnotesize (a) Utility: convert temperature to Kelvin}]
def semantic_function(ctx):
    """Convert a temperature value from Celsius
    or Kelvin to Kelvin."""

    temp_with_unit = load(ctx,
        "the temperature value with unit",
        d_type=str)
    narrate(ctx,
        f"Received temperature: {temp_with_unit}")

    temp_value = extract(ctx, temp_with_unit,
        "the numeric temperature value",
        d_type=float)
    narrate(ctx,
        f"Numeric value: {temp_value}")

    temp_unit = lm(ctx,
        f"What is the temperature unit in "
        f"'{temp_with_unit}'? "
        f"Reply with exactly one of: C, K")
    temp_unit = temp_unit.strip().upper()
    narrate(ctx,
        f"Detected unit: {temp_unit}")

    if "C" in temp_unit:
        temp_kelvin = temp_value + 273
        narrate(ctx,
            f"{temp_value} C + 273 = {temp_kelvin} K")
    else:
        temp_kelvin = temp_value
        narrate(ctx,
            f"Already in Kelvin: {temp_kelvin} K")

    return {
        "the temperature in Kelvin": temp_kelvin
    }
\end{lstlisting}
\caption{Semantic functions constituting the agentic reasoning cache entry for the question in \Cref{fig:example_question_geom}.
Function~(a) is a utility that converts temperature to Kelvin; it uses \texttt{extract()} to obtain the numeric value and \texttt{lm()} to classify the unit. Both of these require an LLM call at invocation time.
Function~(b) on the next page is the root function that orchestrates the full ideal-gas-law computation.
A third function (not shown) converts pressure from kPa to MPa, bar, and atm via pure arithmetic using \texttt{load()} and \texttt{narrate()}.
Runtime primitives \texttt{load}, \texttt{do}, \texttt{narrate}, \texttt{extract}, and \texttt{lm} are highlighted.}
\label{fig:example_sf_geom}
\end{figure}

\begin{lstlisting}[style=python, basicstyle=\ttfamily\footnotesize, title={\footnotesize (b) Root: compute ideal gas pressure}]
def semantic_function(ctx):
    """Calculate the pressure exerted by a gas
    using PV = nRT, rearranged to P = nRT/V."""

    molar_amount_str = load(ctx,
        "the molar amount of gas with unit",
        d_type=str)
    volume_str = load(ctx,
        "the volume of the container with unit",
        d_type=str)
    temperature_str = load(ctx,
        "the temperature with unit", d_type=str)
    gas_name = load(ctx,
        "the common name of the gas", d_type=str)
    gas_model = load(ctx,
        "the gas model or assumption to use",
        d_type=str)

    narrate(ctx, f"Calculating pressure for "
            f"{gas_name} using the {gas_model} law")
    narrate(ctx, f"Given: n = {molar_amount_str}, "
            f"V = {volume_str}, T = {temperature_str}")

    n = extract(ctx, molar_amount_str,
        "the numeric molar amount", d_type=float)
    volume_value = extract(ctx, volume_str,
        "the numeric volume value", d_type=float)

    # Convert temperature via utility function
    temp_result = do(ctx,
        "Convert a temperature value from Celsius "
        "or Kelvin to Kelvin given a temperature "
        "string with unit",
        payloads={
            "the temperature value with unit":
                temperature_str})
    T = load(temp_result,
        "the temperature in Kelvin",
        d_type=float, ctx=ctx)

    R = 8.314  # kPa.dm^3.mol^-1.K^-1
    narrate(ctx, f"P = nRT/V = {n} * {R} * {T}"
            f" / {volume_value}")

    pressure_kpa = n * R * T / volume_value
    narrate(ctx, f"P = {pressure_kpa:.2f} kPa")

    # Convert to other units via utility function
    pressure_result = do(ctx,
        "Convert a pressure value from kPa to "
        "multiple common units including MPa, "
        "bar, and atm",
        payloads={
            "the pressure value in kilopascals "
            "(kPa)": pressure_kpa})

    pressure_mpa = load(pressure_result,
        "the pressure in MPa",
        d_type=float, ctx=ctx)

    return {
        "the calculated pressure in kPa":
            pressure_kpa,
        "the calculated pressure in MPa":
            pressure_mpa,
        # ... bar, atm, summary omitted ...
    }
\end{lstlisting}
\captionof{figure}{Root semantic function of the cache entry (continued from \Cref{fig:example_sf_geom}).
It loads five parameters from the context via \texttt{load()}, extracts numeric values with \texttt{extract()}, delegates temperature conversion and pressure unit conversion to utility functions via \texttt{do()}, retrieves results via \texttt{load(result,\,\ldots)}, and performs the arithmetic $P = nRT/V$.
For brevity, volume-unit handling, additional unit outputs, and some \texttt{narrate()} calls are omitted; in the actual cache entry, every loaded value is narrated.}
\label{fig:example_sf_root_geom}

\begin{lstlisting}[style=python, basicstyle=\ttfamily\footnotesize, title={\footnotesize Compiled semantic function network (names abbreviated; \texttt{narrate()} calls preserved but omitted here)}]
def convert_temperature(ctx,
    temp_with_unit: str,
) -> 'dict':
    temp_value = extract(ctx, temp_with_unit,
        "the numeric temperature value",
        d_type=float)
    temp_unit = lm(ctx,
        f"What is the temperature unit in "
        f"'{temp_with_unit}'? "
        f"Reply with exactly one of: C, K")
    temp_unit = temp_unit.strip().upper()
    if "C" in temp_unit:
        temp_kelvin = temp_value + 273
    else:
        temp_kelvin = temp_value
    return {"the temperature in Kelvin": temp_kelvin}


def convert_pressure(ctx,
    pressure_kpa: float,
) -> 'dict':
    return {
        "the pressure in kPa": pressure_kpa,
        "the pressure in MPa": pressure_kpa / 1000,
        "the pressure in bar": pressure_kpa / 100,
        "the pressure in atm": pressure_kpa / 101.325,
    }


def compute_ideal_gas_pressure(ctx,
    molar_amount: str,
    volume: str,
    temperature: str,
    gas_name: str,
    gas_model: str,
) -> 'dict':
    n = extract(ctx, molar_amount,
        "the numeric molar amount", d_type=float)
    V = extract(ctx, volume,
        "the numeric volume value", d_type=float)
    temp_result = convert_temperature(ctx,
        temp_with_unit=temperature)
    T = temp_result["the temperature in Kelvin"]
    R = 8.314
    pressure_kpa = n * R * T / V
    pres_result = convert_pressure(ctx,
        pressure_kpa=pressure_kpa)
    return {
        "the calculated pressure in kPa":
            pressure_kpa,
        "the calculated pressure in MPa":
            pres_result["the pressure in MPa"],
        "the calculated pressure in bar":
            pres_result["the pressure in bar"],
        "the calculated pressure in atm":
            pres_result["the pressure in atm"],
    }
\end{lstlisting}
\captionof{figure}{Compiled version of the semantic functions from \Cref{fig:example_sf_geom}.
The compiler has statically resolved \texttt{load(ctx,\,\ldots)} calls into named parameters, \texttt{do()} dispatches into direct function calls (lines~41, 46), and \texttt{load(result,\,\ldots)} into dictionary key accesses (line~43).
Crucially, the \texttt{lm()} call (line~7) and \texttt{extract()} calls (lines~4, 37, 39) \emph{cannot} be statically resolved because they require an LLM call at invocation time, and thus survive compilation unchanged.
The pressure-conversion utility compiles to pure arithmetic with no LLM dependency.}
\label{fig:example_compiled_geom}

\begin{figure}[H]
\begin{mdframed}[backgroundcolor=codebg, linecolor=codegray]
{\small \textbf{Output dictionary:}}
\begin{lstlisting}[style=python, basicstyle=\ttfamily\footnotesize, frame=none, numbers=none, backgroundcolor=\color{codebg}]
{'the calculated pressure in kPa': 5132.10,
 'the calculated pressure in MPa': 5.13,
 'the calculated pressure in bar': 51.32,
 'the calculated pressure in atm': 50.65,
 'summary of the calculation':
    'The pressure exerted by 10.0 mol of ethane '
    'in 4.86 dm^3 at 300.0 K is 5.13 MPa '
    '(or 5.13 x 10^3 kPa)'}
\end{lstlisting}
\end{mdframed}

\medskip

\begin{mdframed}[backgroundcolor=codebg, linecolor=codegray]
{\small \textbf{Narration trace} (curated from \texttt{narrate()} calls):}

{\footnotesize\ttfamily
\begin{tabbing}
\hspace{2em}\=\kill
Calculating pressure for ethane using the perfect gas law\\
Given: n = 10.0 mol, V = 4.860 dm\^{}3, T = 27$^{\circ}$C\\
Extracted amount of substance: n = 10.0 mol\\
Extracted volume: V = 4.86 dm\^{}3\\
Converting temperature\ldots\\
\> Received temperature specification: 27$^{\circ}$C\\
\> Extracted numeric temperature value: 27.0\\
\> Detected temperature unit: C\\
\> 27.0$^{\circ}$C + 273 = 300.0 K\\
Temperature in Kelvin: T = 300.0 K\\
Using R = 8.314 kPa$\cdot$dm\^{}3$\cdot$mol\^{}-1$\cdot$K\^{}-1\\
P = nRT/V = 10.0 $\times$ 8.314 $\times$ 300.0 / 4.86\\
P = 5132.10 kPa\\
Converting pressure to other units\ldots\\
\> 5132.10 kPa / 1000 = 5.13 MPa\\
\> 5132.10 kPa / 100 = 51.32 bar\\
\> 5132.10 kPa / 101.325 = 50.65 atm\\
Final: 5132.10 kPa = 5.13 MPa = 51.32 bar = 50.65 atm
\end{tabbing}
}
\end{mdframed}
\caption{Output and execution narration of the cache entry from \Cref{fig:example_sf_geom} on the original question.
The narration traces each reasoning step performed by the semantic functions; indented lines show steps from sub-functions invoked via \texttt{do()}.
The output $\approx 5130\;\mathrm{kPa}$ matches both the source model's answer (\Cref{fig:example_transcript_geom}) and the golden label.}
\label{fig:example_output_geom}
\end{figure}

\section{Worked example: thermodynamics}
\label{appendix:worked_example_thermo}

This appendix walks through a second benchmark problem end-to-end, complementing the ideal gas law example in \Cref{appendix:worked_example_geom}.
The example is a university-level thermodynamics problem from SciBench~\citep{wang2023scibench} (constant-volume work calculation) with two semantic functions.
Both the source model's answer and the cache output match the golden label ($w = 0$).
This example was selected because it is short, scientifically accessible, and exercises all core mechanisms: parameter extraction via \texttt{load()}, sub-function delegation via \texttt{do()}, reasoning narration via \texttt{narrate()}, and value extraction via \texttt{extract()}.

\begin{figure}[H]
\begin{mdframed}[backgroundcolor=codebg, linecolor=codegray]
Assume all gases are perfect unless stated otherwise. Note that 1\,atm = 1.013\,25\,bar. Unless otherwise stated, thermochemical data are for 298.15\,K.
A sample consisting of $2.0\;\mathrm{mol}\;\mathrm{CO_2}$ occupies a fixed volume of $15.0\;\mathrm{dm^3}$ at $300\;\mathrm{K}$.
When it is supplied with $2.35\;\mathrm{kJ}$ of energy as heat its temperature increases to $341\;\mathrm{K}$.
Assume that $\mathrm{CO_2}$ is described by the van der Waals equation of state, and calculate~$w$.
\end{mdframed}
\caption{Original benchmark question from SciBench (university-level thermodynamics).}
\label{fig:example_question_thermo}
\end{figure}

\begin{figure}[H]
\begin{mdframed}[backgroundcolor=codebg, linecolor=codegray]
{\small
\textbf{Calculating work for a constant volume process}

\medskip
\textbf{Given information.}\quad
$n = 2.0\;\text{mol CO}_2$,\;
$V = 15.0\;\text{dm}^3$ (fixed/constant),\;
$T_1 = 300\;\text{K} \to T_2 = 341\;\text{K}$,\;
$q = {+}2.35\;\text{kJ}$.

\medskip
\textbf{Analysis.}\quad
The key phrase is ``fixed volume'', which indicates a constant volume (isochoric) process.

\medskip
\textbf{Calculating work.}\quad
Work done by a gas during expansion or compression is defined as
$w = -\!\int P_{\text{ext}}\,dV$.
For a constant volume process, $dV = 0$, therefore
$w = -\!\int P_{\text{ext}} \cdot 0 = 0$.

\medskip
\textbf{Important note.}\quad
The equation of state (whether ideal gas or van der Waals) describes the \textbf{relationship between $P$, $V$, and $T$} but does not change the fundamental definition of mechanical work.
Since the volume remains constant throughout the process, \textbf{no expansion work is performed regardless of which equation of state the gas follows}.

\medskip
\textbf{Answer.}\quad $\boxed{w = 0}$
}
\end{mdframed}
\caption{Source model $\mathcal{S}$ reasoning transcript for the question in \Cref{fig:example_question_thermo}. The answer $w = 0$ matches the golden label.}
\label{fig:example_transcript_thermo}
\end{figure}

\begin{figure}[H]
\begin{lstlisting}[style=python, basicstyle=\ttfamily\footnotesize, literate={~=}{{$\approx$}}2, title={\footnotesize (a) Utility: classify thermodynamic process type}]
def semantic_function(ctx):
    """Classify the type of thermodynamic process
    based on constraints on state variables."""

    # Load the volume constraint
    volume_constraint = load(ctx,
        "constraint on the volume during the process",
        d_type=str)
    narrate(ctx, f"Volume constraint specified: "
            f"'{volume_constraint}'")

    # Load temperature values
    initial_temp = load(ctx,
        "initial temperature of the gas with units",
        d_type=str)
    final_temp = load(ctx,
        "final temperature of the gas with units",
        d_type=str)

    # Determine if volume is constant
    volume_constant_keywords = [
        "fixed", "constant", "unchanging", "unchanged"]
    is_volume_constant = any(
        kw in volume_constraint.lower()
        for kw in volume_constant_keywords)
    narrate(ctx, f"Volume is constant: {is_volume_constant}")

    # Extract numeric temperatures to check if isothermal
    initial_temp_value = extract(ctx, initial_temp,
        "the numeric temperature value, "
        "e.g. 300 or 273.15", d_type=float)
    final_temp_value = extract(ctx, final_temp,
        "the numeric temperature value, "
        "e.g. 341 or 350", d_type=float)
    is_temperature_constant = (
        abs(initial_temp_value - final_temp_value) < 0.01)

    # Classify the process type
    if is_volume_constant:
        process_type = "isochoric"
    elif is_temperature_constant:
        process_type = "isothermal"
    else:
        process_type = "general"
    narrate(ctx, f"Process classified as {process_type}")

    return {
        "process type": process_type,
        "is volume constant": is_volume_constant,
        "is temperature constant": is_temperature_constant,
    }
\end{lstlisting}
\caption{Semantic functions constituting the agentic reasoning cache entry for the question in \Cref{fig:example_question_thermo}.
Function~(a) is a utility that classifies the thermodynamic process; function~(b) on the next page is the root function that orchestrates parameter loading, delegates to~(a) via \texttt{do()}, and computes the result.
Runtime primitives \texttt{load}, \texttt{do}, \texttt{narrate}, and \texttt{extract} are highlighted.}
\label{fig:example_sf_thermo}
\end{figure}

\begin{lstlisting}[style=python, basicstyle=\ttfamily\footnotesize, title={\footnotesize (b) Root: calculate work done by a gas}]
def semantic_function(ctx):
    """Calculate work done by/on a gas during a
    thermodynamic process given system parameters
    and process constraints."""

    narrate(ctx, "Loading thermodynamic system parameters...")
    amount = load(ctx,
        "amount of substance of the gas sample with units",
        d_type=str)
    narrate(ctx, f"Amount of substance: {amount}")
    gas_identity = load(ctx,
        "chemical identity of the gas", d_type=str)
    volume_constraint = load(ctx,
        "constraint on the volume during the process",
        d_type=str)
    volume = load(ctx,
        "volume of the gas sample with units", d_type=str)
    initial_temp = load(ctx,
        "initial temperature of the gas with units",
        d_type=str)
    final_temp = load(ctx,
        "final temperature of the gas with units",
        d_type=str)
    heat_supplied = load(ctx,
        "heat energy supplied to the system with units",
        d_type=str)
    equation_of_state = load(ctx,
        "equation of state governing the gas behavior",
        d_type=str)
    quantity_to_calculate = load(ctx,
        "thermodynamic quantity to be calculated",
        d_type=str)
    # ... narrate() for each parameter omitted ...

    # Classify the thermodynamic process
    narrate(ctx, "Classifying the thermodynamic process "
            "based on constraints...")
    classification_result = do(ctx,
        "Classify the thermodynamic process type "
        "based on volume and temperature constraints,"
        " determining if the process is isochoric, "
        "isothermal, or general",
        payloads={
            "constraint on the volume during the "
            "process": volume_constraint,
            "initial temperature of the gas with "
            "units": initial_temp,
            "final temperature of the gas with "
            "units": final_temp,
        })

    process_type = load(classification_result,
        "process type", d_type=str, ctx=ctx)
    is_volume_constant = load(classification_result,
        "is volume constant", d_type=bool, ctx=ctx)

    # Calculate work based on process type
    if process_type == "isochoric" or is_volume_constant:
        narrate(ctx, f"Volume is {volume_constraint} "
                f"(constant), dV = 0")
        narrate(ctx, "w = -integral(P_ext * 0) = 0")
        work_value = 0
        work_with_units = "0"
        narrate(ctx, f"Result is independent of "
                f"{equation_of_state}")
        explanation = (
            "For an isochoric (constant volume) "
            "process, w = 0 because dV = 0. "
            "This result is independent of the "
            f"equation of state ({equation_of_state}).")
    else:
        work_value = None
        work_with_units = "requires further calculation"
        explanation = (
            f"For a {process_type} process, work "
            "must be calculated by integration.")

    narrate(ctx, f"Final result: work w = {work_value}")
    return {
        "work value": work_value,
        "work with units": work_with_units,
        "explanation": explanation,
    }
\end{lstlisting}
\captionof{figure}{Root semantic function of the cache entry (continued from \Cref{fig:example_sf_thermo}).
It loads nine parameters from the context via \texttt{load()}, delegates process classification to function~(a) via \texttt{do()}, retrieves the classification result via \texttt{load(result, ...)}, and branches on the process type.
For brevity, individual \texttt{narrate()} calls after each parameter load are omitted; in the actual cache entry, every loaded value is narrated.}
\label{fig:example_sf_root_thermo}

\begin{lstlisting}[style=python, basicstyle=\ttfamily\footnotesize, title={\footnotesize Compiled semantic function network (names abbreviated; \texttt{narrate()} calls preserved but omitted here)}]
def classify_process(ctx,
    volume_constraint: str,
    initial_temperature: str,
    final_temperature: str,
) -> 'dict':
    volume_constraint_ = convert(ctx, volume_constraint, str)
    initial_temp = convert(ctx, initial_temperature, str)
    final_temp = convert(ctx, final_temperature, str)
    volume_constant_keywords = [
        'fixed', 'constant', 'unchanging', 'unchanged']
    is_volume_constant = any(
        kw in volume_constraint_.lower()
        for kw in volume_constant_keywords)
    initial_temp_value = extract(ctx, initial_temp,
        'the numeric temperature value, '
        'e.g. 300 or 273.15', d_type=float)
    final_temp_value = extract(ctx, final_temp,
        'the numeric temperature value, '
        'e.g. 341 or 350', d_type=float)
    is_temperature_constant = (
        abs(initial_temp_value - final_temp_value) < 0.01)
    if is_volume_constant:
        process_type = 'isochoric'
    elif is_temperature_constant:
        process_type = 'isothermal'
    else:
        process_type = 'general'
    return {'process type': process_type,
            'is volume constant': is_volume_constant,
            'is temperature constant':
                is_temperature_constant}


def calculate_work(ctx,
    substance_amount: str, gas_identity: str,
    volume_constraint: str, volume: str,
    initial_temperature: str,
    final_temperature: str,
    heat_supplied: str, equation_of_state: str,
    quantity_to_calculate: str,
) -> 'dict':
    # ... parameter convert() and narrate() calls ...
    classification = classify_process(ctx,
        volume_constraint=volume_constraint,
        initial_temperature=initial_temperature,
        final_temperature=final_temperature)
    process_type = convert(ctx,
        classification['process type'], str)
    is_volume_constant = convert(ctx,
        classification['is volume constant'], bool)
    if process_type == 'isochoric' or is_volume_constant:
        work_value = 0
        work_with_units = '0'
        explanation = ('For an isochoric process, '
            f'w = 0. Independent of '
            f'{equation_of_state}.')
    else:
        work_value = None
        work_with_units = 'requires further calculation'
        explanation = (f'For a {process_type} process, '
            'work must be calculated by integration.')
    return {'work value': work_value,
            'work with units': work_with_units,
            'explanation': explanation}
\end{lstlisting}
\captionof{figure}{Compiled version of the semantic functions from \Cref{fig:example_sf_thermo}.
The compiler has statically resolved all \texttt{load(ctx,\,...)} calls into named parameters, all \texttt{do()} calls into direct function invocations (line~44), and all \texttt{load(result,\,...)} dictionary lookups into concrete key accesses (lines~48--50).
No embedding engine or vector store is required at runtime.
Function and parameter names are abbreviated for readability; the compiler derives them from the natural-language descriptions by sanitization.}
\label{fig:example_compiled_thermo}

\begin{figure}[H]
\begin{mdframed}[backgroundcolor=codebg, linecolor=codegray]
{\small \textbf{Output dictionary:}}
\begin{lstlisting}[style=python, basicstyle=\ttfamily\footnotesize, frame=none, numbers=none, backgroundcolor=\color{codebg}]
{'work value': 0,
 'work with units': '0',
 'explanation': 'For an isochoric (constant volume) '
    'process, w = 0 because dV = 0. This result is '
    'independent of the equation of state '
    '(van der Waals).'}
\end{lstlisting}
\end{mdframed}

\medskip

\begin{mdframed}[backgroundcolor=codebg, linecolor=codegray]
{\small \textbf{Narration trace} (curated from \texttt{narrate()} calls):}

{\footnotesize\ttfamily
\begin{tabbing}
\hspace{2em}\=\kill
Loading thermodynamic system parameters\ldots\\
Amount of substance: 2.0 mol\\
Gas identity: CO\_2\\
Volume constraint: fixed\\
Volume: 15.0 dm\^{}3\\
Initial temperature: 300 K\\
Final temperature: 341 K\\
Heat energy supplied: 2.35 kJ\\
Equation of state: van der Waals\\
Quantity to calculate: w\\
Classifying the thermodynamic process\ldots\\
\> Volume constraint specified: `fixed'\\
\> Volume is constant: True\\
\> Process classified as ISOCHORIC\\
Process classification: isochoric, volume constant: True\\
Volume is fixed (constant), dV = 0\\
w = $-\!\int P_\text{ext} \cdot 0 = 0$\\
Result is independent of equation of state\\
Final result: work w = 0
\end{tabbing}
}
\end{mdframed}
\caption{Output and execution narration of the cache entry from \Cref{fig:example_sf_thermo} on the original question.
The narration traces each reasoning step performed by the semantic functions; indented lines show steps from the utility function invoked via \texttt{do()}.
The output $w = 0$ matches both the source model's answer (\Cref{fig:example_transcript_thermo}) and the golden label.}
\label{fig:example_output_thermo}
\end{figure}

\clearpage
\section{Detailed cost breakdowns}
\label{appendix:cost_breakdowns}

This appendix provides per-role token usage and cost breakdowns for each experiment with available data.
In each table, \textit{Avg~In} and \textit{Avg~Out} are mean input and output tokens per sample;
\textit{Cache~Rd} is the mean number of prompt-cache tokens reused (read) per sample;
\textit{Cache~Wr} is the mean number of tokens written to the prompt cache per sample (applicable only to the builder's multi-turn loop);
and \textit{Cost} is the mean per-sample dollar cost for that role.
Source averages are divided by the total number of samples; builder averages by the number of newly built cache entries; target, extractor, and converter averages by the total number of samples.

\begin{table}[H]
\centering
\footnotesize
\caption{MATH-500.}
\label{tab:cost_math500}
\begin{tabular}{l rrrrr}
\toprule
Role & Avg In & Avg Out & Cache Rd & Cache Wr & Cost (\$) \\
\midrule
Source    & 114     & 1,529   & 0       & ---      & 0.0388 \\
Builder   & 355,140 & 27,591  & 324,207 & 41,587   & 1.1818 \\
Target    & 546     & 393     & 0       & ---      & 0.0001 \\
Extractor & 5,271   & 369     & 0       & ---      & 0.0002 \\
Converter & 1,732   & 596     & 0       & ---      & 0.0002 \\
\midrule
\multicolumn{5}{r}{Total} & 1.2210 \\
\bottomrule
\end{tabular}
\end{table}

\begin{table}[H]
\centering
\footnotesize
\caption{AIME~'24.}
\label{tab:cost_aime24}
\begin{tabular}{l rrrrr}
\toprule
Role & Avg In & Avg Out & Cache Rd & Cache Wr & Cost (\$) \\
\midrule
Source    & 150     & 4,374   & 0       & ---      & 0.1101 \\
Builder   & 451,054 & 32,829  & 414,183 & 49,435   & 1.4198 \\
Target    & 226     & 76      & 0       & ---      & 0.0000 \\
Extractor & 5,851   & 383     & 0       & ---      & 0.0002 \\
Converter & 2,419   & 697     & 0       & ---      & 0.0002 \\
\midrule
\multicolumn{5}{r}{Total} & 1.5303 \\
\bottomrule
\end{tabular}
\end{table}

\begin{table}[H]
\centering
\footnotesize
\caption{AIME~'25.}
\label{tab:cost_aime25}
\begin{tabular}{l rrrrr}
\toprule
Role & Avg In & Avg Out & Cache Rd & Cache Wr & Cost (\$) \\
\midrule
Source    & 225     & 4,766   & 0       & ---      & 0.1203 \\
Builder   & 439,165 & 33,702  & 400,518 & 51,325   & 1.4517 \\
Target    & 494     & 205     & 0       & ---      & 0.0001 \\
Extractor & 8,369   & 461     & 0       & ---      & 0.0003 \\
Converter & 1,231   & 268     & 0       & ---      & 0.0001 \\
\midrule
\multicolumn{5}{r}{Total} & 1.5724 \\
\bottomrule
\end{tabular}
\end{table}

\begin{table}[H]
\centering
\footnotesize
\caption{AIME~'26.}
\label{tab:cost_aime26}
\begin{tabular}{l rrrrr}
\toprule
Role & Avg In & Avg Out & Cache Rd & Cache Wr & Cost (\$) \\
\midrule
Source    & 168     & 4,754   & 0       & ---      & 0.1197 \\
Builder   & 462,836 & 36,763  & 424,685 & 53,344   & 1.5477 \\
Target    & 289     & 116     & 0       & ---      & 0.0000 \\
Extractor & 7,232   & 393     & 0       & ---      & 0.0002 \\
Converter & 1,996   & 555     & 0       & ---      & 0.0002 \\
\midrule
\multicolumn{5}{r}{Total} & 1.6678 \\
\bottomrule
\end{tabular}
\end{table}

\begin{table}[H]
\centering
\footnotesize
\caption{GPQA Diamond.}
\label{tab:cost_gpqa}
\begin{tabular}{l rrrrr}
\toprule
Role & Avg In & Avg Out & Cache Rd & Cache Wr & Cost (\$) \\
\midrule
Source    & 268     & 2,779   & 0       & ---      & 0.0708 \\
Builder   & 597,730 & 41,200  & 553,857 & 60,194   & 1.7781 \\
Target    & 4,796   & 4,519   & 0       & ---      & 0.0010 \\
Extractor & 18,175  & 1,400   & 0       & ---      & 0.0007 \\
Converter & 1,729   & 729     & 0       & ---      & 0.0002 \\
\midrule
\multicolumn{5}{r}{Total} & 1.8508 \\
\bottomrule
\end{tabular}
\end{table}

\begin{table}[H]
\centering
\footnotesize
\caption{GSM8K.}
\label{tab:cost_gsm8k}
\begin{tabular}{l rrrrr}
\toprule
Role & Avg In & Avg Out & Cache Rd & Cache Wr & Cost (\$) \\
\midrule
Source    & 100     & 386     & 0       & ---      & 0.0101 \\
Builder   & 164,598 & 21,110  & 140,887 & 31,599   & 0.8481 \\
Target    & 317     & 96      & 0       & ---      & 0.0000 \\
Extractor & 6,458   & 415     & 0       & ---      & 0.0002 \\
Converter & 4,174   & 1,404   & 0       & ---      & 0.0004 \\
\midrule
\multicolumn{5}{r}{Total} & 0.8589 \\
\bottomrule
\end{tabular}
\end{table}

\begin{table}[H]
\centering
\footnotesize
\caption{MMLU-Pro.}
\label{tab:cost_mmlupro}
\begin{tabular}{l rrrrr}
\toprule
Role & Avg In & Avg Out & Cache Rd & Cache Wr & Cost (\$) \\
\midrule
Source    & 246     & 1,332   & 0       & ---      & 0.0345 \\
Builder   & 477,206 & 32,045  & 439,344 & 48,698   & 1.4084 \\
Target    & 3,707   & 3,035   & 0       & ---      & 0.0007 \\
Extractor & 14,786  & 1,449   & 0       & ---      & 0.0006 \\
Converter & 2,074   & 863     & 0       & ---      & 0.0002 \\
\midrule
\multicolumn{5}{r}{Total} & 1.4445 \\
\bottomrule
\end{tabular}
\end{table}

\begin{table}[H]
\centering
\footnotesize
\caption{SciBench.}
\label{tab:cost_scibench}
\begin{tabular}{l rrrrr}
\toprule
Role & Avg In & Avg Out & Cache Rd & Cache Wr & Cost (\$) \\
\midrule
Source    & 130     & 1,709   & 0       & ---      & 0.0434 \\
Builder   & 293,107 & 30,957  & 261,293 & 45,101   & 1.2572 \\
Target    & 736     & 397     & 0       & ---      & 0.0001 \\
Extractor & 7,655   & 518     & 0       & ---      & 0.0003 \\
Converter & 1,513   & 607     & 0       & ---      & 0.0002 \\
\midrule
\multicolumn{5}{r}{Total} & 1.3011 \\
\bottomrule
\end{tabular}
\end{table}

\clearpage
\section{Target model production}
\label{appendix:target_model_production}

This appendix details the three-stage pipeline used to produce the finetuned 1B-parameter target model $\mathcal{T}$ discussed in \Cref{subsection:target_model_choice}: collecting real invocation logs, augmenting them with synthetic data, and supervised fine-tuning.

\subsection{Stage 1: Collecting invocation logs}

During ordinary cache construction and evaluation runs the system logs every LM call made by semantic functions.
Each log record contains the \emph{role} of the caller (builder, extractor, converter, or target), the full request prompt, and the model's response.
We assign every record a deterministic identifier derived from the experiment name, sample id, perturbation tag, role, and call index to enable deduplication and reproducibility.

After running the evaluation benchmarks described in \Cref{section:evaluation}, we harvest roughly 20\% of the resulting logs.
Only records whose role is relevant to the target model (i.e.\ the target, extractor, and converter roles) are retained, as the builder and source roles exercise capabilities beyond what $\mathcal{T}$ needs.
The filtered records are serialized as a HuggingFace \texttt{Dataset} on disk, forming the \emph{seeding} dataset for the next stage.

\subsection{Stage 2: Synthetic data augmentation}

A seeding set of a few thousand real invocations is too small to finetune a language model reliably.
We therefore amplify it with synthetic variants, targeting a roughly $10\times$ expansion.

For each seeding record, we repeatedly prompt the source model $\mathcal{S}$ (Claude Opus 4.5) to generate three synthetic variants at a time, aiming for ten valid variants in total.
Each variant is a structurally identical request string that differs only in its \emph{payload} (i.e., the concrete question, problem statement, or data being processed) while preserving the instructional scaffold verbatim.
The prompt wraps the seeding request and its response (for context) in XML tags and instructs the model to emit each variant inside \texttt{<variant>} tags.
A regex parser extracts the variants and discards degenerate outputs shorter than ten characters or containing leaked chain-of-thought.
Within each round, if fewer than three valid variants are recovered, the generator retries up to twice with an appropriately reduced request count.

Once all variant requests have been generated, a second pass through the synthetically generated variants fills in the corresponding responses.
Each synthetic request is sent to $\mathcal{S}$ and the reply is recorded alongside the request.
The full set of synthetic (request, response) pairs is written to batched JSONL files and finally consolidated into a single HuggingFace \texttt{Dataset}.

\subsection{Stage 3: Supervised fine-tuning}

The consolidated dataset is used to finetune \textbf{Llama~3.2~1B}~\cite{meta2024llama32} with supervised fine-tuning (SFT) via the \texttt{trl} library's \texttt{SFTTrainer}.
Each training example is formatted as a single-turn conversation: the request becomes a \texttt{user} message and the response an \texttt{assistant} message.
Rows with empty completions are discarded.

\Cref{tab:finetune_hyperparams} lists the hyperparameters.
Training uses BFloat16 mixed precision, gradient checkpointing for memory efficiency, and the AdamW optimizer.
The trainer evaluates validation loss at the end of every epoch and saves a checkpoint; early stopping with a patience of three epochs halts training when validation loss ceases to improve, and the best checkpoint is restored automatically.
The final model and tokenizer are saved as a self-contained artifact that can be used on an edge device or served through an inference endpoint.

\begin{table}[h]
\centering
\small
\caption{Fine-tuning hyperparameters for the 1B target model.}
\label{tab:finetune_hyperparams}
\begin{tabular}{ll}
\toprule
Parameter & Value \\
\midrule
Base model            & \texttt{meta-llama/Llama-3.2-1B} \\
Max training epochs   & 20 \\
Learning rate         & $2 \times 10^{-5}$ \\
Optimizer             & AdamW \\
Per-device batch size  & 8 (auto-scaled) \\
Gradient accumulation & 1 \\
Max sequence length   & 4\,096 tokens \\
Precision             & BFloat16 \\
Gradient checkpointing & Enabled \\
Evaluation strategy   & Every epoch \\
Early stopping patience & 3 epochs \\
Best-model metric     & Validation loss \\
\bottomrule
\end{tabular}
\end{table}

Multi-GPU training is supported via HuggingFace \texttt{accelerate}; in practice a single GPU sufficed for the 1B-parameter model given the moderate dataset size.

\end{document}